\documentclass[final]{cvpr}

\usepackage{times}
\usepackage{epsfig}
\usepackage{graphicx}
\usepackage{amsmath}
\usepackage{amssymb}
\usepackage{booktabs}

\newcommand*\rot{\rotatebox{90}}

\newcommand{\bfsection}[1]{\vspace*{0.1cm}\noindent\textbf{#1.}}

\usepackage[pagebackref=true,breaklinks=true,colorlinks,bookmarks=false]{hyperref}

\def\cvprPaperID{6} 
\def\confYear{CVPR 2021}

\begin{document}

\title{Object Propagation via Inter-Frame Attentions for Temporally Stable Video Instance Segmentation}

\author{Anirudh S Chakravarthy\textsuperscript{1,2}
\and Won-Dong Jang\textsuperscript{2} \and Zudi Lin\textsuperscript{2} \and Donglai Wei\textsuperscript{2} \and Song Bai\textsuperscript{3} \and Hanspeter Pfister\textsuperscript{2} \and\\
\textsuperscript{1}Birla Institute of Technology and Science, Pilani, India\\
\textsuperscript{2}School of Engineering and Applied Sciences, Harvard University, Cambridge, USA\\
\textsuperscript{3}University of Oxford, UK}

\maketitle

\begin{abstract}
Video instance segmentation aims to detect, segment, and track objects in a video. Current approaches extend image-level segmentation algorithms to the temporal domain. However, this results in temporally inconsistent masks.
In this work, we identify the mask quality due to temporal stability as a performance bottleneck. Motivated by this, we propose a video instance segmentation method that alleviates the problem due to missing detections. Since this cannot be solved simply using spatial information, we leverage temporal context using inter-frame attentions. This allows our network to refocus on missing objects using box predictions from the neighbouring frame, thereby overcoming missing detections.
Our method significantly outperforms previous state-of-the-art algorithms using the Mask R-CNN backbone, by achieving 36.0\% mAP on the YouTube-VIS benchmark. Additionally, our method is completely online and requires no future frames. Our code is publicly available at \url{https://github.com/anirudh-chakravarthy/ObjProp}.
\end{abstract}

\section{Introduction}




\begin{figure}[t]
    \centering
    \includegraphics[width=0.32\columnwidth]{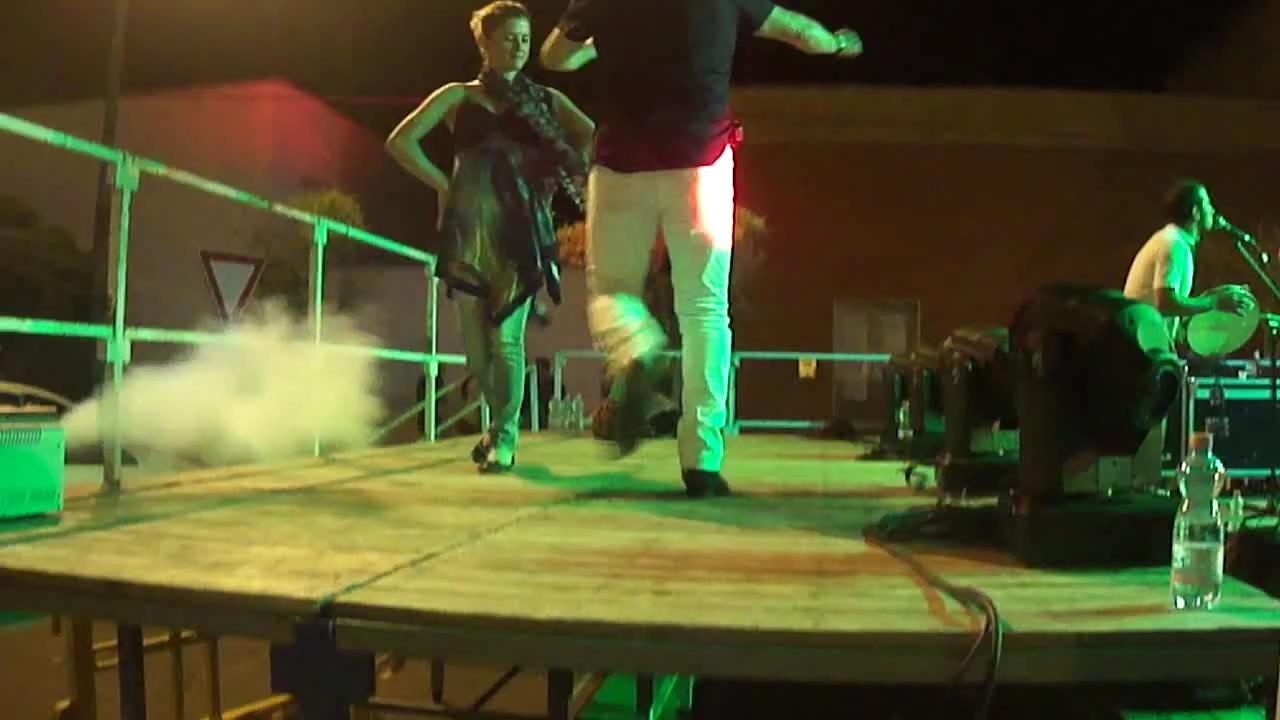}
    \includegraphics[width=0.32\columnwidth]{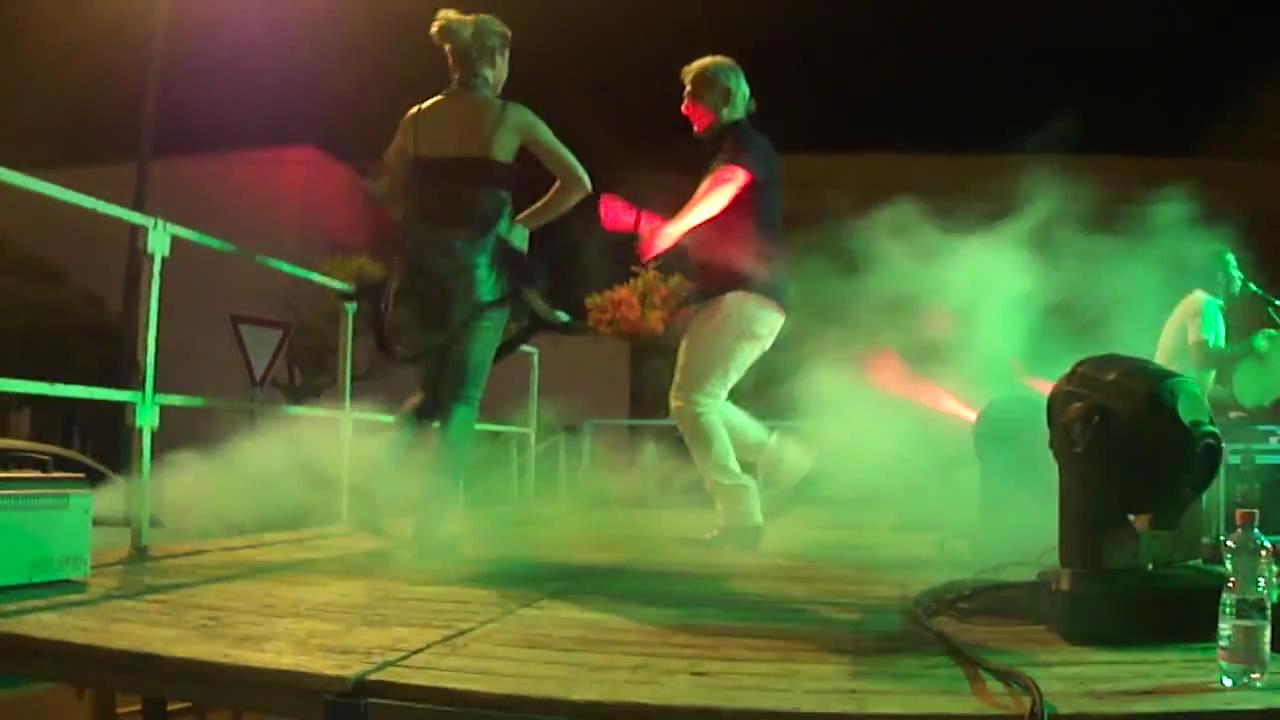}
    \includegraphics[width=0.32\columnwidth]{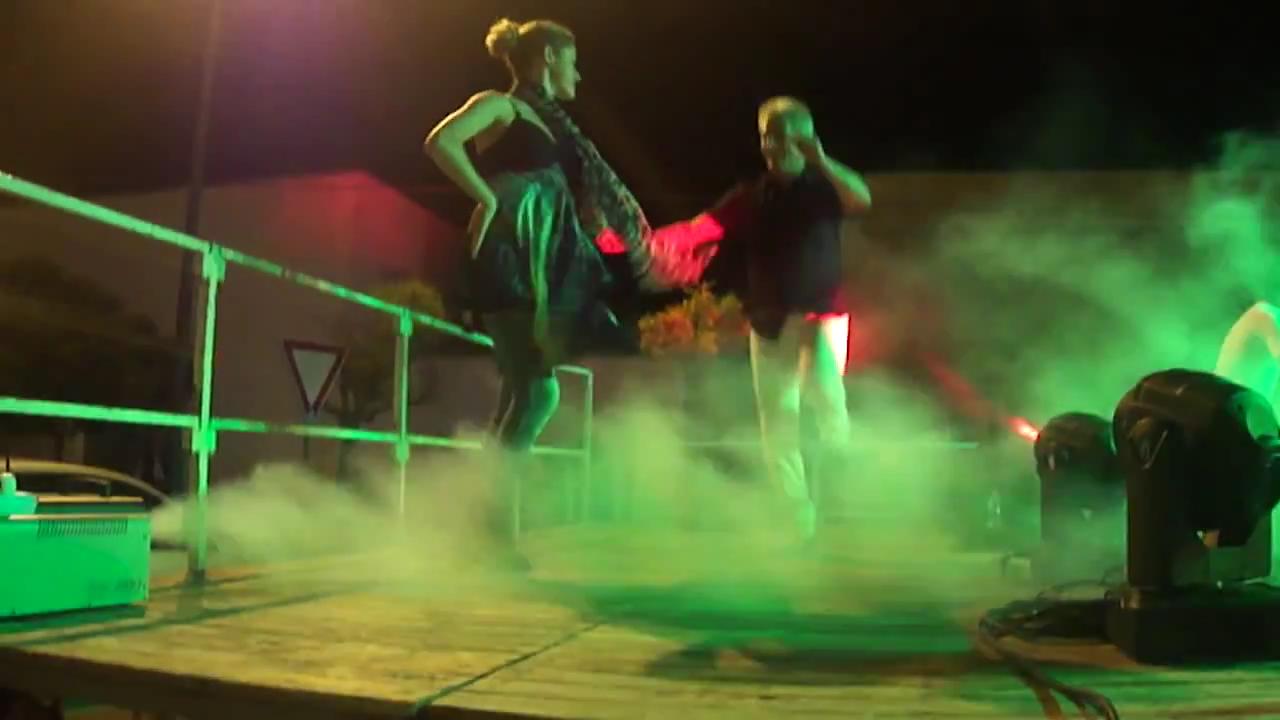}
    \includegraphics[width=0.32\columnwidth]{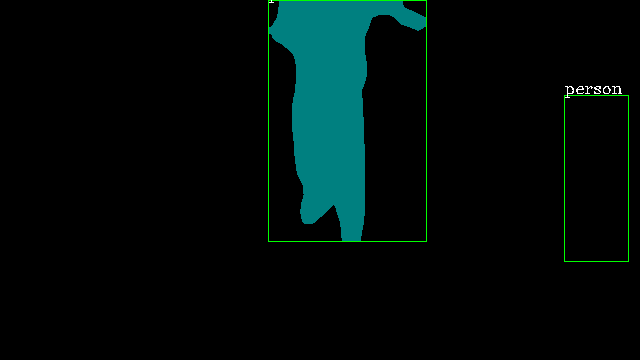}
    \includegraphics[width=0.32\columnwidth]{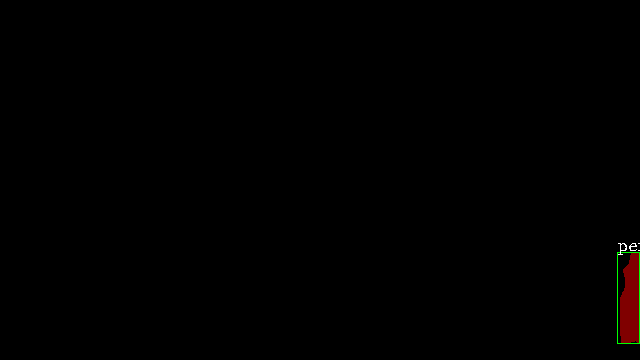}
    \includegraphics[width=0.32\columnwidth]{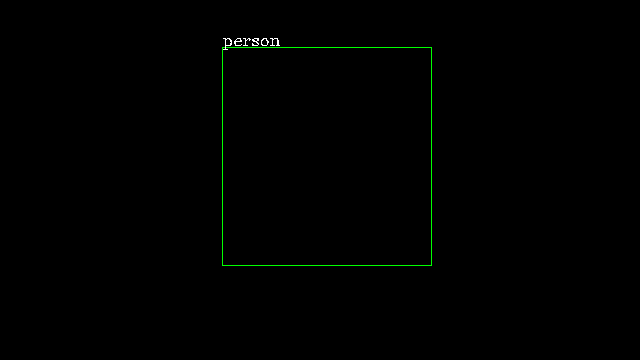}
    \includegraphics[width=0.32\columnwidth]{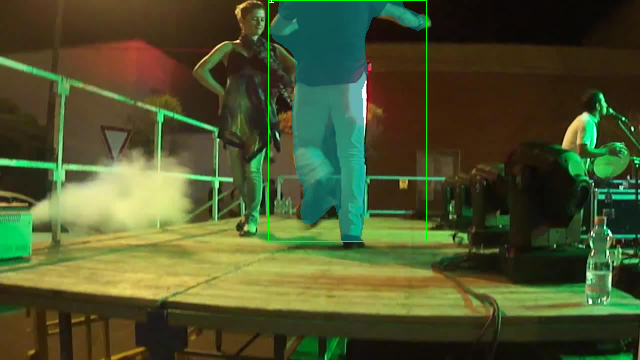}
    \includegraphics[width=0.32\columnwidth]{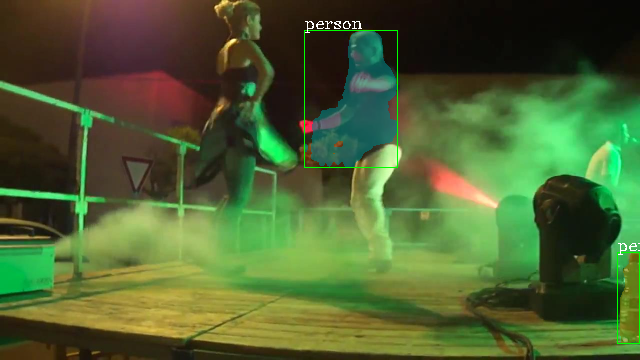}
    \includegraphics[width=0.32\columnwidth]{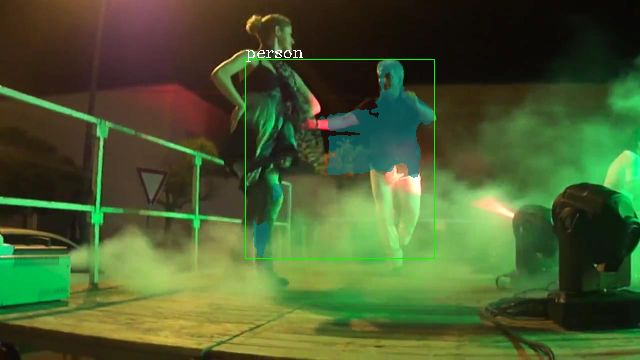}
    \caption{We demonstrate the problem of temporal stability in video instance segmentation. Due to object displacements, mask and class predictions are temporally inconsistent in MaskTrack R-CNN~\cite{yang2019Video} (second row). Our method alleviates the issue of temporal stability (third row).
    }
    \label{fig:teaser}
\end{figure}



In this work, we propose a video instance segmentation algorithm based on the Mask R-CNN pipeline~\cite{he2017Mask}.
We focus on the problem of temporal instability in video instance segmentation (Figure~\ref{fig:teaser}). There are many reasons for temporal instability: missing proposals from a region proposal network, misclassification of the object's class, or aliasing from small visual displacements. 
We address temporal instability by propagating masks using object boxes to neighbouring frames to complement missing detections. 
Mask propagation through bounding boxes enables tracking of objects even when the detector misses the object's bounding box in the current frame.
To this end, we use the attention mechanism. Our propagation network predicts an attention map propagated from previous frames to the current frame. We apply the attention map to current frame features, which allows us to fill in absent instance masks and overcome temporal instability. 

We have three main contributions in this work. First, we identify the temporal instability for video instance segmentation. Second, by propagating object masks through an inter-frame attention mechanism, we generate temporally coherent and spatially accurate mask tracks. Third, our method outperforms the conventional Mask R-CNN methods on the on the YouTube-VIS dataset~\cite{yang2019Video}.

\begin{figure*}[t]
    \centering
    \includegraphics[width=0.8\textwidth]{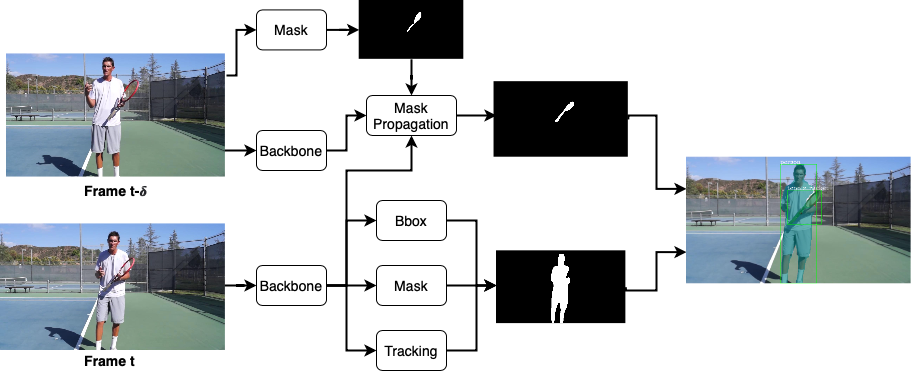}
    \caption{We add the mask propagation branch to MaskTrack R-CNN~\cite{yang2019Video}. Given two frames from a video, we propagate an attention map from frame $t-\delta$ (randomly sampled) to frame $t$ to predict segmentation masks corresponding to missing instances. 
    }
    \label{fig:pipeline}
\end{figure*}

\section{Method}\label{sec:method}

\begin{table}[t]
    \caption{Oracle study using MaskTrack R-CNN~\cite{yang2019Video} on our mini-validation set from the YouTube-VIS dataset. The highest gain achieved by correcting masks is {\bf highligted}.
    }\label{table:oracle}
    \begin{center}
    \begin{tabular}{cccccc}
    \toprule 
     Box & Class & Mask & Track & mAP & Gain\\
     \midrule
     -- & -- & -- & -- & 36.8 & --\\
     \checkmark & -- & -- & -- & 41.9 & +5.1\\
     \checkmark & \checkmark & -- & -- & 53.1 & +11.2\\
     \checkmark & \checkmark & \checkmark & -- & 85.3 & \textbf{+32.2}\\
     \checkmark & \checkmark & \checkmark & \checkmark & 100.0 & +14.7\\
    \bottomrule
    \end{tabular}
    \end{center}
\end{table}


\subsection{Oracle study}
In order to examine the problem of temporal stability in greater detail, we perform an oracle testing in Table~\ref{table:oracle} using MaskTrack R-CNN~\cite{yang2019Video}.
MaskTrack R-CNN achieves 36.8 mAP on the mini-validation set. Replacing the detected boxes with the closest ground-truth boxes based on intersection over union (IoU) achieves a performance of 41.9 mAP. Completely replacing the box head with ground truth boxes and category labels leads to an improvement of 11.2 mAP. When we replace the mask head with ground truth masks, a significant performance improvement is observed from 53.1 mAP to 85.3 mAP. Furthermore, the oracle tracking improves the performance by 14.7 points. 

Motivated by this study, we found that the temporal instability of masks is the critical bottleneck, which means there is considerable room for improvement in mask generation and tracking. 
Thus, we focus on improving the quality of masks, especially alleviating the temporal stability problem in video instance segmentation.

\subsection{Mask Propagation with Inter-frame Attentions}
\begin{figure*}[t]
    \centering
    \includegraphics[width=0.8\textwidth]{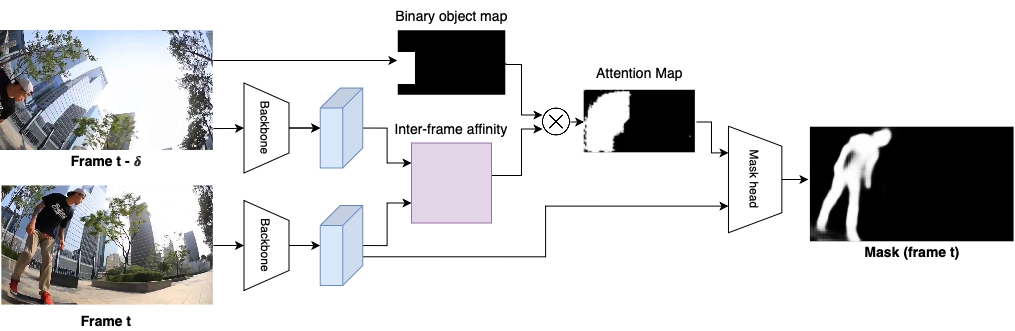}
    \caption{An illustration of our attention-guided mask propagation system which propagates instance masks from frame $t-\delta$ to frame $t$. Our framework can be summarized in three steps. 1) An affinity matrix is computed between the two frames. 2) Next, the instance-specific box-level mask and affinity matrix are used to propagate an attention map. 3) The attention map is applied on frame $t$ and the segmentation mask is predicted using a mask prediction head. 
    }
    \label{fig:prop_head}
\end{figure*}

Due to object deformation or aliasing, per-frame object instance segmentation models struggle to segment objects consistently throughout the video. This leads to several missing detections throughout the video, which drastically affects the segmentation results. 
Using 3D convolutional neural networks (CNNs) poses a couple of challenges. First, 3D convolutions are computationally expensive. Second, since the large memory footprint constrains the number of images in memory, the learned temporal interaction is limited.

Instead, we propagate masks from previous frames $t-\delta$ to the current frame $t$ to compensate the missing detections due to the lack of temporal context. As illustrated in Figure~\ref{fig:pipeline}, we add our propagation module upon the MaskTrack R-CNN pipeline. Our propagation module enables learning of the temporal context without 3D convolutions.
In this work, we improve the transition-based propagation method using attention mechanism~\cite{vaswani2017Attention}.
Our inter-frame attentions can robustly propagate masks between frames considering the temporal context. 

Our propagation module with inter-frame attentions is illustrated in Figure~\ref{fig:prop_head}. For object propagation, we set the backbone features of frames $t-\delta$ and $t$ and a binary map of an object to propagate at frame $t-\delta$ as input. Our propagation module will output a mask of the target object at frame $t$.

\bfsection{Inter-frame affinity}
We first measure inter-frame affinities between two frames $t-\delta$ and $t$. 
The input backbone features from frames $t-\delta$ and $t$ are resized into the stride of 16 of the image resolution and then concatenated across each level of the feature pyramid. We represent these processed features as $\mathbf{F}_t$ and $\mathbf{F}_{t-\delta}$ for frames $t$ and $t-\delta$, respectively. By using these features, we compute the transition matrix to measure inter-frame feature affinity between each spatial location. 
The inter-frame affinity matrix $\mathbf{W}_{t-\delta \rightarrow t} \in \mathbb{R}^{HW \times HW}$ is computed as
\begin{equation}
    \mathbf{W}_{t-\delta \rightarrow t} = \mathbf{F}_t \circ \mathbf{F}_{t-\delta},
\label{eq: affinity}
\end{equation}
where $\circ$ is a matrix multiplication operator. Each element of the inter-frame affinity matrix represents the affinity between corresponding two locations in $t$ and $t-\delta$ frames. We normalize the affinity matrix to make the sum of each row to be 1.

\bfsection{Attention estimation via object propagation}
Similar to TVOS~\cite{zhang2020Transductive}, we aim to propagate masks using the transition matrix. However, instead of propagating pixel-level segmentation masks, we use a binary object map, which serves as a loose estimate for the pixel-level mask. 
The input binary object map for frame $t-\delta$ is generated by marking all pixels within the instance-specific bounding box. In addition to the object map, we also generate a binary map for the background by inverting the object's binary map to take into account the background information when computing inter-frame attentions.
We vectorize both binary object and background maps for matrix multiplication.
Using the transition matrix, the binary object map, $\mathbf{b}_{t-\delta}$, is propagated to the current frame $t$ by
\begin{equation}
    \mathbf{a}_{t} =  \mathbf{W}_{t-\delta \rightarrow t}
    \circ \mathbf{b}_{t-\delta}.
\end{equation}
Since we propagate both object and background maps, we apply softmax to find the regions of the object. We represent the softmax output of the propagated object map as $\hat{\mathbf{a}}_{t}$ and use it as an attention map.
Unlike existing attention-based algorithms which allow the machine to prioritize important features, we explicitly supervise the attention module to learn where to focus by propagating temporal information. This allows us to maximize the temporal context.

\bfsection{Attention-based mask prediction}
Our next aim is to predict a mask using the attention map, $\hat{\mathbf{a}}_{t}$, and the features from frame $t$, $\mathbf{F}_{t}$. We first apply the attention map to the features by conducting element-wise multiplication to each spatial location. We feed the attention-guided features to a mask prediction module, which consists of four convolution modules (a combination of a convolution layer and ReLU), one deconvolution module, and prediction module (a combination of a convolution layer and a sigmoid layer). 



\bfsection{Loss functions}
The propagation loss $L_{prop}$ consists of two terms, the mask propagation loss $L_{prop}^{m}$ and the attention loss $L_{prop}^{a}$. $L_{prop}^{m}$ is computed identically to the mask head in Mask R-CNN~\cite{he2017Mask}. 
The attention loss, $L_{prop}^{a}$, is computed as follows:
\begin{equation}
    L_{prop}^{a} = -\sum_{i=1}^H\sum_{j=1}^W y_{ij} y_{ij} \log \Tilde{y}_{ij} + (1 - y_{ij}) \log (1 - \Tilde{y}_{ij}),
\end{equation}
where $y_{ij}$ is the value in the ground-truth attention map at location $(i, j)$, and $\Tilde{y}_{ij}$ is the predicted attention value.


\bfsection{Training} 
We use Mask R-CNN~\cite{he2017Mask} with ResNet-50 backbone pre-trained on COCO dataset. Our model is trained on 4 Tesla V100 GPUs. We use a batch size of 24, with 6 images on each GPU. We train our model for 12 epochs using SGD optimizer with a learning rate of 0.005, which is decayed by a factor of 10 at 8 and 11 epochs. 

\bfsection{Inference}
During inference, our framework is completely online and does not require any future frames. We store the output predictions for previous frames in memory and propagate instance masks to the current frames. Using mask propagation, we are able to alleviate the temporal instability problem. When a bounding box is missing from the current frame, we propagate an instance mask from the frame history to the current frame to segment the missing object instance. Therefore, we use mask propagation as an empty instance filling mechanism.

\section{Experiments}

\begin{figure*}[t]
    \centering
    \setlength\tabcolsep{1.0pt}
    \begin{tabular}{ccccc}
        \small
        \rot{Video clip} &    \includegraphics[width=0.19\textwidth]{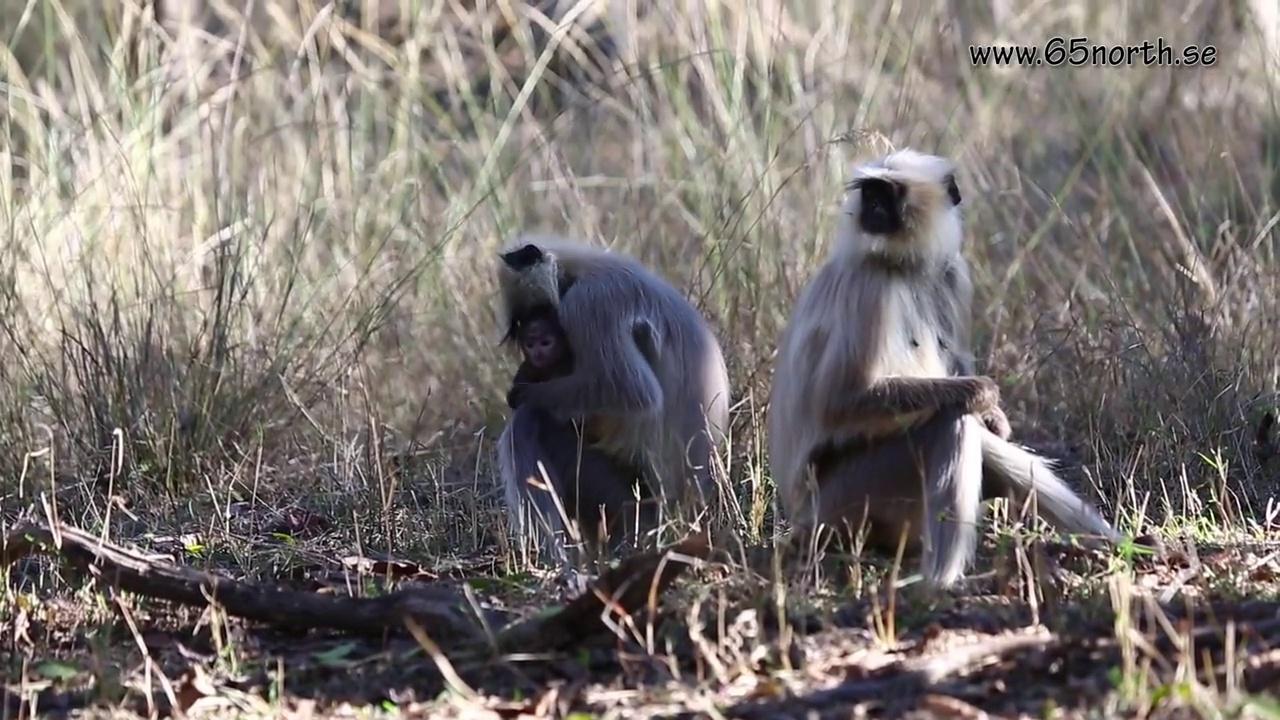} & \includegraphics[width=0.19\textwidth]{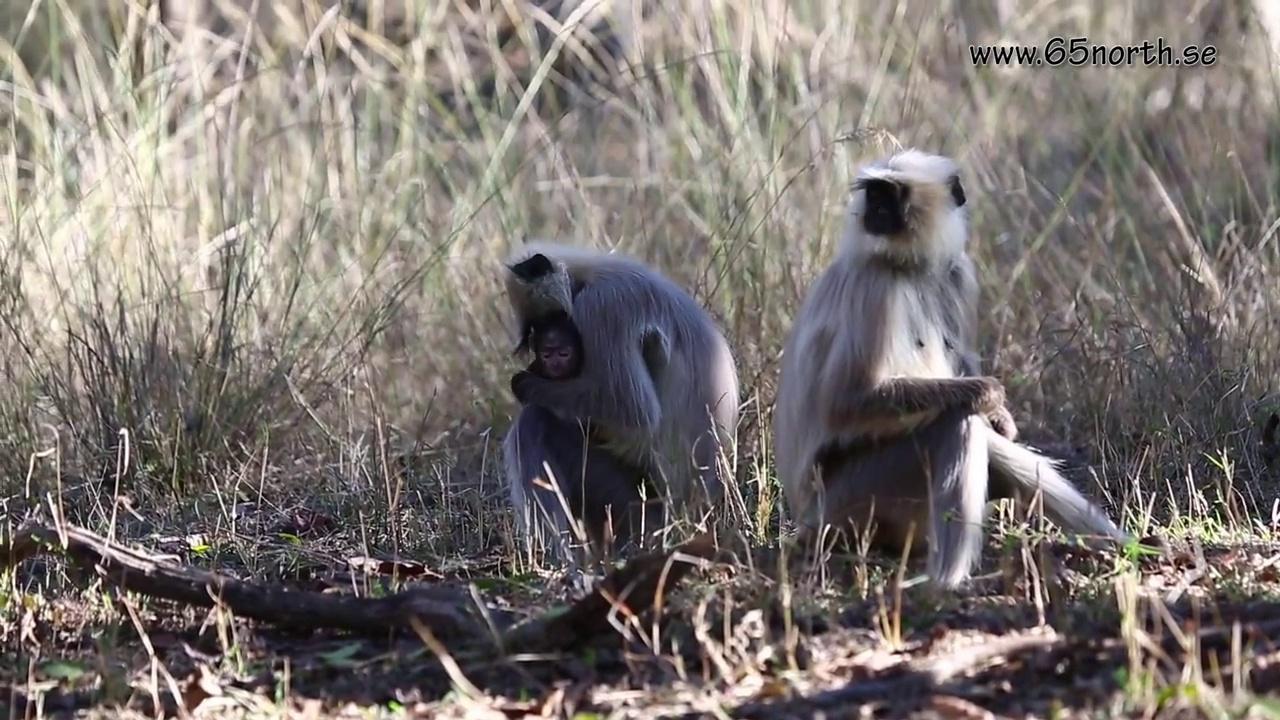} & \includegraphics[width=0.19\textwidth]{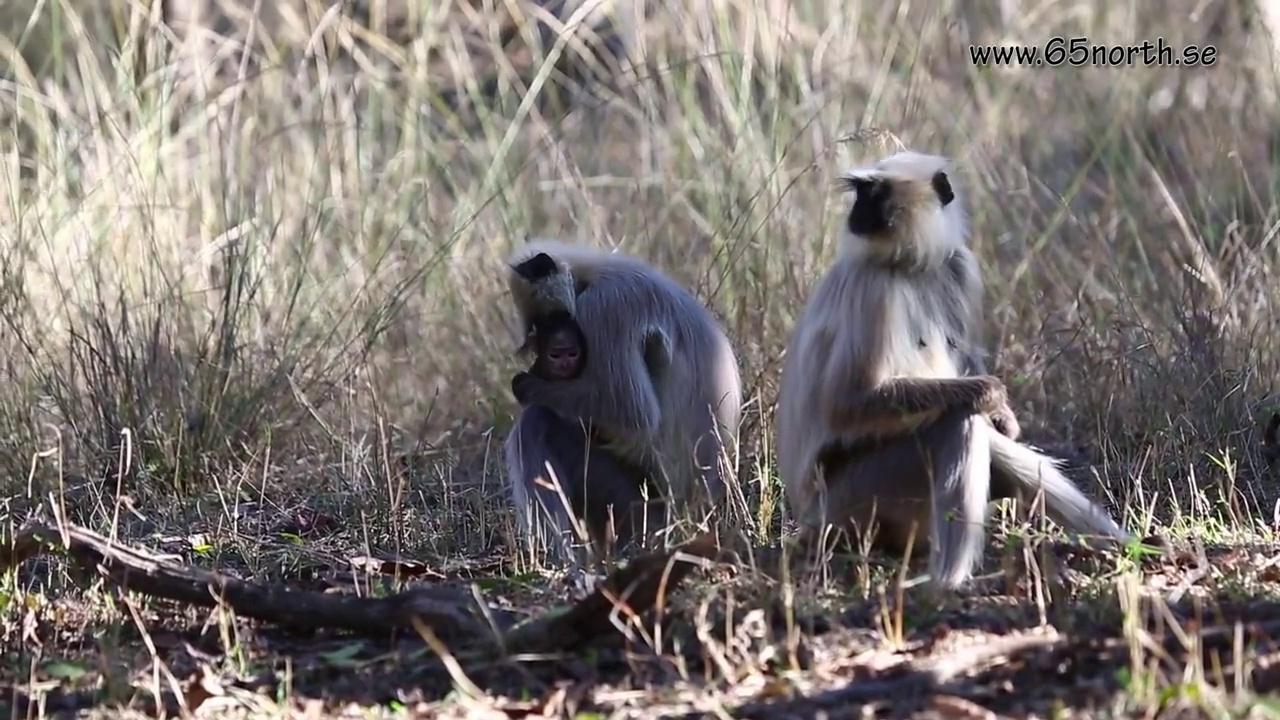} & \includegraphics[width=0.19\textwidth]{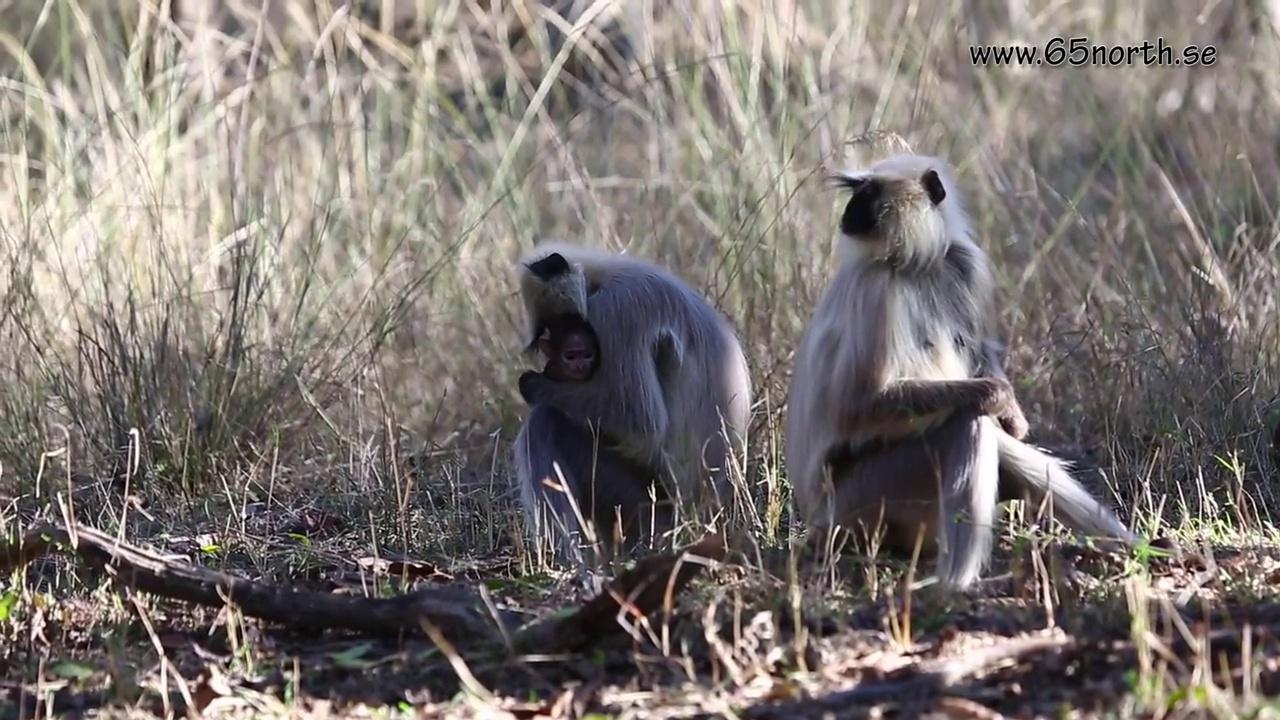}\\
        \small
        \rot{MaskTrack} \rot{R-CNN} & \includegraphics[width=0.19\textwidth]{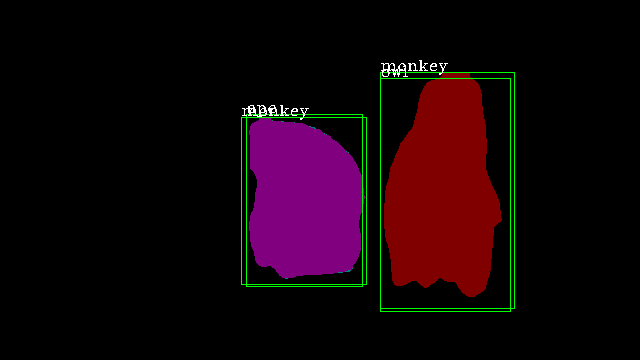} & \includegraphics[width=0.19\textwidth]{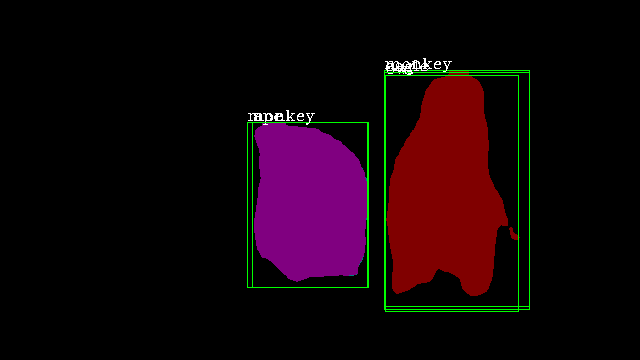} & \includegraphics[width=0.19\textwidth]{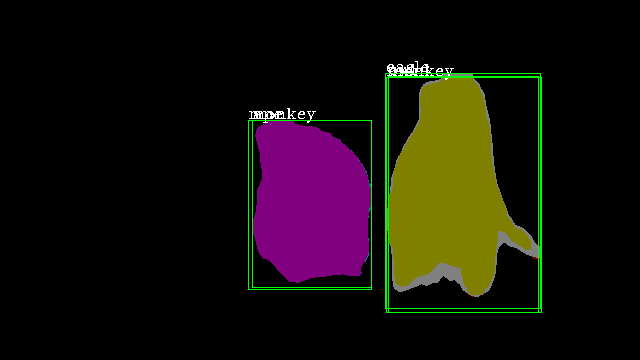} & \includegraphics[width=0.19\textwidth]{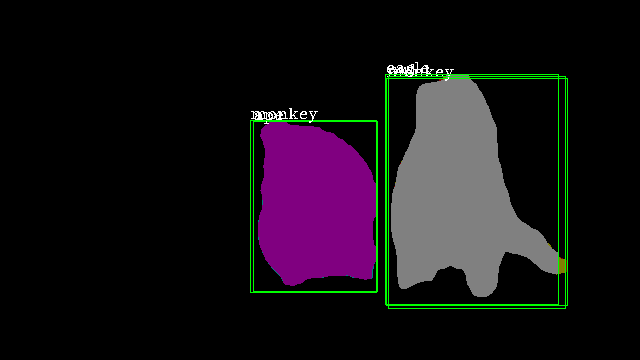}\\
        \small
        \rot{Ours} &
        \includegraphics[width=0.19\textwidth]{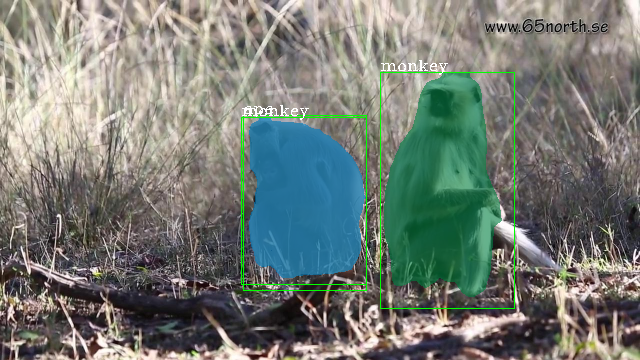} & \includegraphics[width=0.19\textwidth]{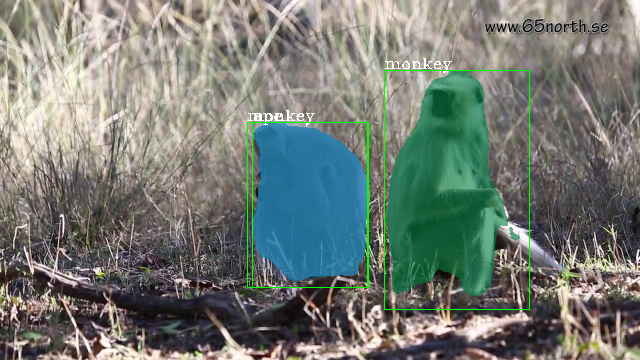} & \includegraphics[width=0.19\textwidth]{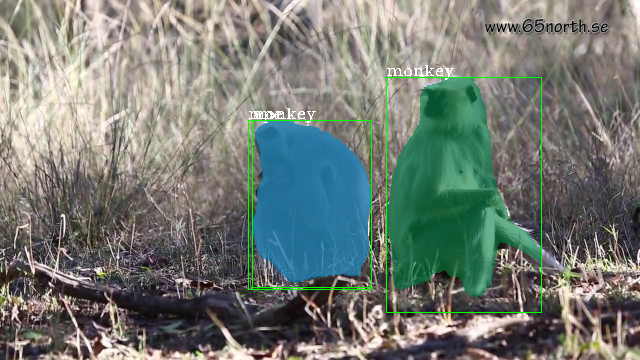} & \includegraphics[width=0.19\textwidth]{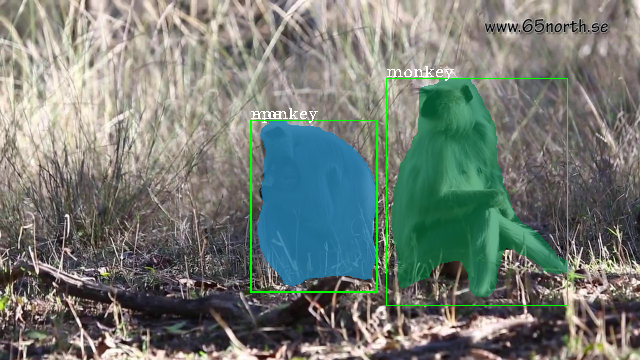} \\
        & Frame 6 & Frame 8 & Frame 9 & Frame 11
    \end{tabular}
    \caption{We compare results obtained using our approach (third row) with MaskTrack R-CNN predictions (second row). 
    }
    \label{fig:results}
\end{figure*}
\begin{table*}[t]
    \caption{Comparison of video instance segmentation methods on the YouTube-VIS validation dataset~\cite{yang2019Video}.
    The best results are boldfaced.}
    \begin{center}
    \resizebox{0.65\textwidth}{!}{
    \begin{tabular}{lcccccc}
    \toprule 
    Method & Backbone & mAP & AP@50 & AP@75 & AR@1 & AR@10\\
    \midrule
    MaskTrack R-CNN & ResNet-50 & 30.3 & 51.1 & 32.6 & 31.0 & 35.5\\

    STEm-Seg & ResNet-50 & 30.6 & 50.7 & 33.5 & 31.6 & 37.1\\
    
    
    SipMask & ResNet-50 & 32.5 & 53.0 & 33.3 & 33.5 & 38.9\\

    Ours & ResNet-50 & \textbf{36.0} & \textbf{59.4} & \textbf{39.2} & \textbf{39.1} & \textbf{47.7}\\
    
    \bottomrule
    \end{tabular}
    } 
    \end{center}
\label{table:yt-vis}
\end{table*}

We use the YouTube-VIS~\cite{yang2019Video} validation set to compare video instance segmentation methods.
For evaluation, we follow 3 metrics. We use 1). mean Average Precision over the video sequence (mAP), 2). Average Precision over the video sequence at 50\% and 75\% IOU thresholds, and 3). Average Recall for the highest 1 and 10 ranked instances per video.
We compare our method with three state-of-the-art methods; MaskTrack R-CNN~\cite{yang2019Video}, STEm-Seg~\cite{athar2020STEmSeg}, and SipMask~\cite{cao2020SipMask}. Note that direct comparison with MaskProp~\cite{bertasius2020Classifying} is not fair due to the complex architecture, hence, we compare our method with Mask R-CNN based approaches.



It is observable that our method comprehensively outperforms all four conventional methods on all evaluation metrics. We achieve nearly 3.5\% greater mAP than the closest method on the benchmark, which demonstrates the effectiveness of our approach. 
In Figure~\ref{fig:results}, we illustrate the results of our method (third row) compared to MaskTrack R-CNN (second row). We observe that our inter-frame attention propagation head leads to temporally consistent segmentation tracks throughout the video.




\section{Conclusions}
In this work, we introduce an inter-frame attention propagation network for video instance segmentation. Using box-level instance masks from the frame history, we propagate an attention map onto the current frame which is used to generate an instance-specific segmentation mask. Our method is online and requires limited computational overhead. 
Using inter-frame attentions, we achieve state-of-the-art results on the YouTube-VIS benchmark using the Mask R-CNN pipeline. 
Qualitative results demonstrate the effectiveness of our approach in alleviating missing detections due to temporal stability problem. 

{\small
\bibliographystyle{ieee_fullname}
\bibliography{egbib}
}

\end{document}